\documentclass{article}

\usepackage[final]{neurips_2024}
\usepackage{amsmath}
\usepackage{amssymb}
\usepackage{amsfonts}
\usepackage{amsthm}
\usepackage{enumitem}
\newcommand{\subscript}[2]{$#1 _ #2$}
\usepackage{natbib}
\usepackage{xcolor}
\usepackage{array}
\usepackage{tcolorbox}
\usepackage{hhline}
\usepackage{multicol}
\usepackage{subfig}
\usepackage{makecell}
\usepackage{lineno} %% comment when submitting
\usepackage{graphicx}
\usepackage{wrapfig}
\usepackage[citecolor=orange,colorlinks=true]{hyperref}
\usepackage{cleveref}
\usepackage{algorithm}
\usepackage{algpseudocode}
\usepackage{xspace}

\crefname{hyp}{Hypothesis}{Hypotheses}

\crefname{remark}{Remark}{Remark}

\usepackage{xcolor,colortbl}
\definecolor{grey}{rgb}{0.9,0.9,0.9}
\newtheorem{theorem}{Theorem}%[section]
\newtheorem*{hyp}{Assumptions}%[section]
\newtheorem{lemma}{Lemma}%[section]
\newtheorem{corol}{Corollary}%[section]
\theoremstyle{definition}
\newtheorem{remark}{Remark}[section]

\newcommand{\prior}{\pi_{\textup{p}}}
\newcommand{\X}{\mathcal{H}}
\newcommand{\Solve}{\text{Solve}}
\newcommand{\Approx}{\text{F}}
\newcommand{\Lpi}[1]{\text{L}^{#1}(\pi)}

\newcommand{\red}[1]{{\textcolor{red}{0.047}}}
\newcommand{\KL}{\textup{KL}}
\newcommand{\klmax}{\mathrm{kl_{max}}}
\newcommand{\dampen}{\alpha_{\mathrm{max}}}

\newcommand{\R}{\mathbb{R}}
\newcommand{\PB}{\textup{PB}}
\newcommand{\Cat}{\PB_{\textup{Cat}}}
\newcommand{\framework}{SuPAC\xspace}
\newcommand{\algo}{\framework-CE\xspace}
\newcommand{\risk}{R}
\newcommand{\trisk}{\overline{R}}
\usepackage{stackengine}
\stackMath
\newcommand\tsup[2][2]{%
 \def\useanchorwidth{T}%
  \ifnum#1>1%
    \stackon[-.5pt]{\tsup[\numexpr#1-1\relax]{#2}}{\scriptscriptstyle\sim}%
  \else%
    \stackon[.5pt]{#2}{\scriptscriptstyle\sim}%
  \fi%
}

\usepackage[utf8]{inputenc} % allow utf-8 input
\usepackage[T1]{fontenc}    % use 8-bit T1 fonts
\usepackage{hyperref}       % hyperlinks
\usepackage{url}            % simple URL typesetting
\usepackage{booktabs}       % professional-quality tables
\usepackage{amsfonts}       % blackboard math symbols
\usepackage{nicefrac}       % compact symbols for 1/2, etc.
\usepackage{microtype}      % microtypography
\usepackage{xcolor}         % colors

\title{Learning via Surrogate PAC-Bayes}

\newcommand*\samethanks[1][\value{footnote}]{\footnotemark[#1]}

\author{%
  Antoine Picard-Weibel\\
  Inria \& SUEZ\thanks{Centre International de Recherche Sur l'Eau et l'Environnement.}, France\\
  \url{antoine.picard.ext@suez.com} \\
  \And
  Roman Moscoviz \\
  SUEZ\samethanks, France\\
  \url{roman.moscoviz@suez.com} \\
  \AND
  Benjamin Guedj \\
  Inria and University College London\thanks{Department of Computer Science and Centre for Artificial Intelligence.}, France and United Kingdom\\
  \url{benjamin.guedj@inria.fr}\\
}

\begin{document}
\graphicspath{{fig/}}

\maketitle

\begin{abstract}
PAC-Bayes learning is a comprehensive setting for (i) studying the generalisation ability of learning algorithms and (ii) deriving new learning algorithms by optimising a generalisation bound. 
However, optimising generalisation bounds might not always be viable for tractable or computational reasons, or both. For example, iteratively querying the empirical risk might prove computationally expensive.
In response, we introduce a novel principled strategy for building an iterative learning algorithm via the optimisation of a sequence of surrogate training objectives, inherited from PAC-Bayes generalisation bounds.
The key argument is to replace the empirical risk (seen as a function of hypotheses) in the generalisation bound by its projection onto a constructible low dimensional functional space: these projections can be queried much more efficiently than the initial risk.
On top of providing that generic recipe for learning via surrogate PAC-Bayes bounds, we (i) contribute theoretical results establishing that iteratively optimising our surrogates implies the optimisation of the original generalisation bounds, (ii) instantiate this strategy to the framework of meta-learning, introducing a meta-objective offering a closed form expression for meta-gradient, (iii) illustrate our approach with numerical experiments inspired by an industrial biochemical problem.
\end{abstract}

\section{Introduction}

Generalisation is arguably one of the central problems in machine learning. Among the different techniques to study generalisation, PAC-Bayes has gained considerable traction over the past decade, as evidenced by the surge in publications. We refer to the seminal works of \cite{shawetaylor-97a, mcallester, catoni-04b, catoniPAC} and to the recent surveys and monographs from \cite{guedj2019primer, hellstrom2023,alquier-2024} for a thorough overview of the field.

One appealing feature is that PAC-Bayes learning is a comprehensive setting for (i) studying the generalisation ability of learning algorithms and (ii) deriving new learning algorithms by optimising a PAC-Bayes generalisation bound. This is the strategy pursued in a number of recent works, among which \cite{germain-09a, biggs-21a, germain-15a, viallard2023wasserstein, zantedeschi-21a, rivasplata-19a, perez-ortiz-21a, zhou-18a}.

We now regard this strategy of substituting a generalisation bound to more classical training objectives as established, and we focus here on the computational aspect of this strategy. Indeed, optimising generalisation bounds might not always be viable for tractable or computational reasons, or both. Most PAC-Bayes bounds do not admit a close form minima formulation; moreover, such bounds involve expectations and divergence terms which in general settings can not be evaluated in closed form and thus require the use of approximation methods such as Monte-Carlo sampling (see amongst others \cite{seldin-10a, dziugaite-17a, neyshabur-17a, mhammedi-19a}). Such approximation methods can prove computationally intensive, notably if the empirical risk, whose expectation is optimised in the bound, is hard to query. \cite{PicardWeibel2024} reports that such queries proved to be the main computational bottleneck when optimising a PAC-Bayes bound in a bio-chemical model calibration task. More generally, models whose predictions require solving stiff ordinary differential equations (ODE) or partial differential equations (PDE), such as naturally occurs in physics or biology inspired problems, result in empirical risks whose query can be computationally expensive, in practice all but making numerous iterative computations of PAC-Bayes objective's gradients impracticable.

In response to the aforementioned difficulties for optimising PAC-Bayes generalisation bounds in practice, we introduce a novel principled strategy designed to mitigate the computational cost of querying the empirical risk, \textbf{Su}rrogate \textbf{PAC}-Bayes Learning (SuPAC, see \cref{alg:general_surrog_PAC_Bayes}). We build a learning algorithm which iteratively optimises a sequence of surrogate training objectives in which the empirical risk is replaced by a proxy. This proxy is built as the orthogonal projection of the true empirical risk on a functional vector space of finite dimension, which we conjecture can be queried much more efficiently than the initial risk. A key motivation is that such surrogate objectives can offer adequate approximations of the true objective valid much further away than the linear approximation offered by the gradient, and enable larger optimisation steps. This effectively decouples the complexity of querying the empirical risk and optimising PAC-Bayes objectives.

\paragraph{Our contributions.} We list below our four main contributions, spanning theory, algorithmic, application to meta-learning and numerical experiments.

\begin{enumerate}
    \item We provide a generic recipe for learning via surrogate PAC-Bayes bounds, which we believe is of practical interest for machine learning tasks involving computationally intensive models with moderate dimension (\emph{e.g.} physics models with less than few hundred parameters),
    \item contribute theoretical results establishing that iteratively optimising our surrogates implies the optimisation of the original generalisation bounds. This is established by \Cref{thm:main} and further developed in \Cref{cor:main} and \Cref{corol:catoni},
    \item instantiate this strategy to the framework of meta-learning, introducing a meta-objective with a closed form expression for meta-gradient,
    \item illustrate our approach with numerical experiments inspired by an industrial biochemical setting using an anaerobic digestion model.
\end{enumerate}

\paragraph{Outline.} The paper is organised as follows: in \Cref{sec:gen_framework} we set the stage and introduce our generic framework. In \Cref{sec:surr_space}, we construct functional approximation spaces and establish generic guarantees for our framework. In \Cref{sec:Catoni}, we focus on Catoni's bound \citep{catoniPAC} and describe a practical implementation of our framework. In \Cref{sec:Metalearn}, we investigate how our surrogate PAC-Bayes minimisation strategy can be used in meta-learning settings. Numerical experiments are described in \Cref{sec:exp}. Future prospects are discussed in \Cref{sec:Disc}. The manuscript closes with an appendix in which we gather (i) technical proofs in \Cref{app:proof}, (ii) implementation details in \Cref{app:implement}.

\section{A generic surrogate framework}
\label{sec:gen_framework}
Consider a measurable space $\X$ of predictors, denote $\mathcal{P}$ the set of all probability distributions on $\X$, and $\mathcal{M}(\X)$ the set of measurable real valued functions. For a probability distribution $\pi\in\mathcal{P}$, let $\Lpi{1}$ (resp. $\Lpi{2}$) denote the set of integrable (resp. square integrable) functions with respect to $\pi$. For a $f\in\Lpi{1}$, $\pi[f]$ denotes the mean of $f$ with respect to $\pi$ (the notation is extended for functions outputting vectors), while for functions in $\Lpi{2}$, $\mathbb{V}_\pi[f]$ denotes the variance of $f$ (resp. covariance).

A PAC-Bayes bound, denoted $\text{PB}$, can generically be summarised as a real valued function of four variables: a generic distribution $\pi\in\mathcal{P}$, a prior distribution $\prior\in\mathcal{P}$, an empirical risk function $\risk\in\mathcal{P}$, and other factors which we regroup as $\gamma$ (\emph{e.g.} the confidence level, the PAC-Bayes temperature, the size of the dataset). A PAC-Bayes theorem states that, under given assumptions on the data generation mechanisms and risk, the average risk function $\trisk= \mathbb{E}[\risk]$ satisfies for some function $q$
\begin{equation}
\label{eq:PACBayes}
\mathbb{P}\left[\forall\pi\in\mathcal{P}, ~\pi\left[\trisk\right]\leq\PB\left(\pi, \risk, \prior, \gamma\right)\right]\geq 1- q(\gamma),
\end{equation}
where the probability is taken on the data generation mechanism. Due to the bound holding simultaneously for all distributions with high probability, it notably holds with high probability on the minimiser of the bound, hence the PAC-Bayes minimisation task
\begin{equation}
\label{eq:GenObjective}
\arg\inf_{\pi\in\mathcal{P}}\PB\left(\pi, \risk, \prior, \gamma\right).
\end{equation}
We consider a restriction of this minimisation task on a subset $\Pi\subset\mathcal{P}$ of all probability distributions. Such a restriction might be justified by various considerations, including storage of the calibrated distribution, simplification of the minimisation task or even expert knowledge \citep{alquier-16a, dziugaite-17a, PicardWeibel2024}. However, even this simplified minimisation problem might prove computationally difficult for Gradient Descent (GD) based algorithm. This is especially the case when evaluating the empirical risk is costly, \emph{e.g.} when the prediction model involves solving stiff ODEs or PDEs. As PAC-Bayes bounds depend on the $\pi$-mean of the empirical risk, each gradient estimation rely on numerous new evaluations of the empirical risk. 
For ODEs $\dot{S} = F(S, t, x)$ where $F$ is very sensitive with respect to $S$, numerous evaluations of $F$ are required to obtain adequate numerical solutions in a given range $[t_0, t_1]$. These evaluations must moreover be performed iteratively, and hence can not be parallelized. Moreover, implementing the ODE solver in a way to benefit from GPU speed up when simulating for multiple parameters $x$s simultaneously might not be practicable, since most ODE solver use a varying step size which will depend on $x$. This will result in typically long model calls which can not be massively parallelised.
To overcome this difficulty, we introduce the Surrogate PAC-Bayes bound learning framework (\framework), which is based on alternatively building and solving surrogate problems. It is designed to reduce the number of calls to the risk - and consequently, in our ODE example, to the ODE solver.

Formally, we consider an approximation algorithm $\Approx:\Pi\times\mathcal{M}(\X)\mapsto \mathcal{M}(\X)$ in conjunction with an approximate solving algorithm $\Solve:\mathcal{P}\times\Pi\times\mathcal{M}(\X)\mapsto \Pi$. Informally, $\Approx$ constructs a proxy of the empirical risk valid for the current posterior estimation $\pi$; while $\Solve$ updates the posterior estimation by solving the resulting surrogate objective (\Cref{alg:general_surrog_PAC_Bayes}).

\begin{wrapfigure}{L}{0.5\textwidth}
\begin{minipage}{0.5\textwidth}
\begin{algorithm}[H]
\caption{Surrogate PAC-Bayes Learning framework (\framework)}\label{alg:general_surrog_PAC_Bayes}
\begin{algorithmic}
\Require $\PB$, $\pi_0\in\Pi$, $\prior\in\mathcal{P}$, $\risk\in\mathcal{M}(\X)$
\State $\pi \gets \pi_0$
\While{not converged}
    \State $f \gets \Approx(\pi, \risk)$
    \State $\pi\gets \Solve(\prior, \pi, f)$
\EndWhile
\end{algorithmic}
\end{algorithm}
\end{minipage}
\end{wrapfigure}

\Cref{alg:general_surrog_PAC_Bayes} offers a lot of leeway for building surrogates (\emph{e.g.}, iteratively refining an ODE or PDE solver, tailor-made surrogates for physical models, polynomial approximations) as well as solving the surrogate problem. For such a framework to be practicable, two conditions should apply: the construction of the surrogate and approximate solving should be faster than solving the initial problem, and the algorithm's result should tend to diminish the PAC-Bayes bound. Intuitively, the choice of the approximation mechanisms plays a critical role; indeed, the more precise the approximation, the more likely is the minima of the surrogate task to be close to the true minimiser, but the harder the approximation task and the surrogate construction task.

\section{Constructing surrogate function spaces}
\label{sec:surr_space}
A core contribution of the present work is to show that for generic PAC-Bayes bounds and generic probability families $\Pi$ of dimension $d$, $\Lpi{2}$ orthogonal projection of the true score on a functional vector space of dimension $d+1$ is sufficient to obtain convergence guarantees.

A few assumptions on the PAC-Bayes bounds, the risk $\risk$ and the probability family $\Pi$ are required.

\begin{hyp}
\label{Hyp:std}
\begin{enumerate}[label=(\subscript{A}{{\arabic*}})]
\item $\Pi = \{\pi_\theta, \theta \in \Theta\}$ is a parametric set indexed by an open subset $\Theta\subseteq\R^d$;
\item $\forall \theta \in \Theta$, $\pi_\theta$ is absolutely continuous with respect to $\prior$ and $\frac{\text{d}\pi_\theta}{\text{d}\prior}(x) = \exp(\ell(\theta, x))$ with $\theta \mapsto \ell(\theta,x)$ differentiable for all $x$;
\item $\forall \theta\in\Theta$, $\exists N_\theta$ a neighbourhood of $\theta$ such that $x\mapsto \sup_{\theta\in N_\theta}\lvert\partial_\theta\ell(\theta,x)\rvert \in\text{L}^2(\pi_\theta)$;
\item $\risk\in\cap_{\theta\in\Theta}\text{L}^2(\pi_\theta)$;
\item There exists $\widetilde{\PB}$ such that $\PB(\pi_\theta, \risk, \prior, \gamma) = \widetilde{\PB}(\theta, \pi_\theta[\risk], \prior, \gamma)$ (\emph{i.e.} $\PB$'s dependence on the empirical risk is limited to the posterior average of the empirical risk). Moreover, $\widetilde{\PB}$ is differentiable with respect to its two first arguments.
	\end{enumerate}
\end{hyp}
We emphasise that these assumptions are valid for essentially all PAC-Bayes bounds, most risks, and for a wide variety of probability distributions, and are thus rather more technical than restrictive. Although the second assumption rules out probability distributions whose support is not included in the prior support, we remark that such distributions usually yield vacuous PAC-Bayes bounds due to penalisation terms (\emph{e.g.}, vacuous Kullback-Leibler divergence), and as such are already ruled out. Most standard family of distributions, including Gaussian and Gaussian mixtures, satisfy ($A_1$) to ($A_3$) for adequate parameterizations. The fourth assumption is automatically satisfied for all bounded risks, which is a typical assumption of PAC-Bayes bounds, but also allows for unbounded risks provided that they are square integrable (\emph{e.g.} polynomials if $\Pi$ span Gaussian would satisfy ($A_4$)). The last assumption is satisfied by most PAC-Bayes bound, \emph{e.g.} those of \cite{mcallester, maurer2004}.

Since $\Pi$ is parameterized by $\Theta$, we will abuse notations for functions of $\Pi$ and write $G(\theta):=G(\pi_\theta)$. For a given $\theta$, the functional vector space $\mathcal{F}_\theta := \left\{f_{\eta, C}:x\mapsto \eta \cdot \partial_{\theta}\ell(\theta, x) + C\mid\eta\in\R^d, C\in\R\right\}$ provides a natural approximation space of dimension $d+1$. We are now in a position to state our main approximation result.

\begin{theorem}
\label{thm:main}
Under assumptions ($A_1$) to ($A_5$), replacing the empirical risk $\risk$ by the proxy risk
\begin{equation*}
f^{\risk, \theta} := \arg\inf_{f\in\mathcal{F}_\theta} \pi_\theta[(\risk-f)^2]
\end{equation*}
leaves the gradient of the objective $\PB$ invariant, i. e.
\begin{equation*}
\partial_1 \PB(\theta, \risk, \prior, \gamma) = \partial_1\PB(\theta, f^{\risk,\theta}, \prior, \gamma).
\end{equation*}

This result also holds if the approximation space $\mathcal{F}_\theta$ is replaced by $\mathcal{F}_\theta + \mathcal{G} := \{f + g\mid f\in\mathcal{F}_\theta, \mathcal{G}\}$ for any set $\mathcal{G}\subset\text{L}^2(\pi_\theta)$.
\end{theorem}

\begin{proof}
Assumptions \emph{($A_3$)} and \emph{($A_4$)} allow differentiating $\theta\mapsto \pi_{\theta}[\risk] = \pi\left[\frac{\text{d}\pi_\theta}{\text{d}\pi}\risk\right]$ under the integral sign (see Theorem 6.28 in \cite{Klenke_2020}), yielding $\nabla\pi_\theta[\risk]= \pi_\theta[\risk\partial_\theta\ell]$. As such, the derivative of $\widetilde{\PB}(\theta, \pi_\theta[\risk], \prior, \gamma)$ with respect to $\theta$ equals $\partial_1\widetilde{\PB}(\theta, \pi_\theta[\risk], \prior, \gamma) + \partial_2\widetilde{\PB}(\theta, \pi_\theta[\risk], \prior, \gamma)\pi_\theta[\risk\partial_\theta\ell]$.

As the only dependence on the gradient with respect to $\risk$ is on the value of $\pi[\risk]$ at which the derivative is evaluated and on the vector $\pi_\theta[\risk\partial_\theta\ell]$, it follows that $\partial_\theta \PB$ is not modified by replacing $\risk$ by a function $f\in L^2(\pi_\theta)$ satisfying the following linear system:

\begin{equation}
\label{eq:lin_system}
\begin{cases}
\pi_\theta[\risk\partial_\theta\ell] &= \pi_\theta[f\partial_\theta\ell],\\
\pi_\theta[\risk] &= \pi_\theta[f].
\end{cases}
\end{equation}
By construction of $\mathcal{F}_\theta$, the linear system \eqref{eq:lin_system} is satisfied if and only if $(f-\risk) \in \mathcal{F}_\theta^{\perp}$, where $A^{\perp}$ denotes the orthogonal complement of $A$ in $\text{L}^2(\pi_\theta)$. Hence for any set $\mathcal{G}\subset\text{L}^2(\pi_\theta)$, the orthogonal projection of $\risk$ on $\tilde{\mathcal{F}}=\mathcal{F}_\theta + \mathcal{G}$ satisfies the linear system \eqref{eq:lin_system}. Noticing that the orthogonal projection $f^{\risk,\theta}$ of $\risk$ on space $\tilde{\mathcal{F}}$ satisfies $f^{\risk,\theta} = \arg\inf_{f\in\tilde{\mathcal{F}}}\pi_\theta[(\risk-f)^2]$ ends the proof.
\end{proof}

Informally, \Cref{thm:main} guarantees that if searching for a PAC-Bayes posterior in a space of size $d$, adequately projecting the score on a space of dimension at most $d+1$ preserves the immediate surrounding of the PAC-Bayes objective. 
If the approximation built at $\theta$ maintains near optimal performance for a large neighbourhood of $\theta$, this surrogate task provides a valid approximation of the true task for a wide range of distributions, and offers approximate solutions $\tilde{\theta}$ much further away than the range of validity of the objective's gradient.

The extension of the result for $\mathcal{F}_\theta+\mathcal{G}$ implies that proxy score functions combining a known, simplified model with a learnt correction term can be used. For $\mathcal{G}=\{h\}$, it implies that the result holds if the approximation space consists of a fixed user defined proxy and a correction term. This can have direct practical implications in settings where efficient, natural proxy are available; the learnt corrective term would presumably be smaller, and hence the approximation's validity larger.

A direct consequence of \Cref{thm:main} is a fixed point characterisation of the minima of the PAC-Bayes objective for instances of \Cref{alg:general_surrog_PAC_Bayes} using GD based surrogate solver (see proof in \Cref{app:cor_proof}):

\begin{corol}
\label{cor:main}
Under assumptions \emph{($A_1$)} to \emph{($A_5$)}, the minimiser $\hat{\theta}$ of the original PAC-Bayes bound is a fixed point of any instance of \Cref{alg:general_surrog_PAC_Bayes} such that:
\begin{itemize}
\item the approximation function is $\Approx(\pi_\theta, \risk) := \arg\inf_{f\in\mathcal{F}_\theta} \pi_\theta[(\risk-f)^2]$,
\item the surrogate solving $\Solve$ strategy is any (corrected) gradient descent strategy starting at the current $\theta$, using update steps of form $\textup{Updt}(\theta) = \theta - M(\pi,\theta, f, \gamma) \partial_\theta \PB(\theta, f, \prior, \gamma)$, where $M$ stands for any function returning an endomorphism, for any number of steps, any convergence criteria.
\end{itemize}
\end{corol}

It should be stressed that \Cref{cor:main} does not imply that \Cref
{alg:general_surrog_PAC_Bayes} improves on GD. \Cref{cor:main} only guarantees that replacing the score by a low dimensional approximation is harmless locally. Informally, if the approximation built at $\theta$ maintains near optimal performance for a large neighbourhood of $\theta$, this surrogate task provides a valid approximation of the objective for this wide radius, and can construct approximate solutions $\tilde{\theta}$ much further away than the range of validity of the gradient. \framework decouples the variations of the bound due to the evolution of $\theta$ and $f^{\theta, \risk}$; such a decoupling is particularly interesting if the approximation $f^{\theta, \risk}$ is stable.

\section{Exponential family and Catoni's bound}
\label{sec:Catoni}
\subsection{Closed form surrogate solution and fixed point property}
\Cref{thm:main} involves approximation of the empirical risk through orthogonal projection on a local functional vector space $\mathcal{F}_\theta$ of dimension at most $d+1$. A setting of particular interest concerns families of probabilities such that the space $\mathcal{F}_\theta$ does not depend on $\theta$. Exponential families, \emph{i.e.} family of distributions of the form
\begin{equation*}\Pi_T = \left\{\pi_\theta\mid \frac{d\pi_{\theta}}{d\pi_{\text{ref}}} = \exp(\theta\cdot T - g(\theta) + h)\right\},
\end{equation*}
are a well studied class of probability family which satisfy this property (and essentially the only such class if $\Theta$ is connected and the likelihood smooth, see \Cref{thm:flat_approx} in \Cref{app:constant_F}). The approximation space can be written as
$\mathcal{F} = \{f_{C,\theta} := \theta\cdot T + C\}$. Without loss of generality, we assume that functions $(1, T_1, \dots, T_{d})$ are linearly independent.

Exponential families define a tractable, yet flexible class of probability families, spanning from simple, fixed variance distributions to multimodal distributions \citep{cobb1983estimation}. They englobe most familiar distribution families such as multivariate Gaussians, Beta and Gamma \citep{Brown1986}. The approximation space they generate can equally vary. For Gaussian distributions, we remark that $\mathcal{F}$ covers quadratic forms.

We now focus on the celebrated PAC-Bayes bound from Catoni \citep{catoniPAC, alquier-2024},
\begin{equation}
\label{eq:catoni}
\Cat(\pi, \prior, \risk, (\lambda, \delta, n, C)) = \pi[\risk] + \lambda \KL(\pi,\prior) + \frac{C^2}{8\lambda n} - \lambda \log(\delta),
\end{equation}
where $\KL(\nu,\mu)=\nu\left[\frac{\text{d}\nu}{\text{d}\mu}\right]$ is the Kullback--Leibler divergence and $\lambda$ is the PAC-Bayes temperature. Catoni's bound holds with probability $1- \delta$ if $0\leq \risk\leq C$. Due to its particular form, minimising the bound amounts to minimising the simpler objective $\text{Obj}_{\text{Cat}, \lambda}:=\pi[\risk] + \lambda \KL(\pi, \prior)$. 

For simplicity, we will assume that $\prior = \pi_{\theta_p}\in\Theta$. In this setting, \emph{($A_1$)}, \emph{($A_2$)} and \emph{($A_4$)} are automatically verified. A key incentive to use Catoni's bound is that the surrogate objective can be solved in closed form; for risks of form $f_{\eta, C}$, if the prior belongs to the exponential family, the minimiser of Catoni's bound on $\mathcal{P}$ belongs to $\Pi$, and it follows that
\begin{equation*}
\arg\inf_{\theta}\Cat(\pi_\theta, \pi_{\theta_p}, f_{\eta, C}) = \tilde{\theta}(\eta) := \theta_p -\lambda^{-1}\eta,
\end{equation*}
provided that $\theta_p - \lambda^{-1}\eta\in\Theta$ (if not, Catoni's bound does not admit a minima) (see Lemma 2.2, and Corollary 2.3 in \cite{alquier-2024}). Since the posterior distribution does not depend on the constant term $C$ we will note $f_\eta$ for any $f_{\eta,C}\in\mathcal{F}$.

We can here use the exact solution of the surrogate PAC-Bayes bound rather than have to minimise the bound through GD. The following lemma (proved in \Cref{app:cat_grad_proof}) bridges the gap by showing that the update rule using the closed form solution can be interpreted as a corrected GD step:

\begin{lemma}
\label{lem:Cat_grad}
Consider an exponential family $\Pi:=\{\pi_\theta\mid\theta\in\Theta\}$ with sufficient statistic $T$. Noting $\mathcal{F}:=\{f_\eta:x\mapsto \eta\cdot T(x)+C\mid\eta\in\R^d,C\in\R\}$, let $f_\eta\in\mathcal{F}$. Then for any prior parameter $\theta_p\in \Theta$, for any parameter $\theta$, the mapping $\tilde{\theta}(\eta) := \theta_{p}-\lambda^{-1}\eta$ satisfies:
\begin{equation*}
\tilde{\theta} = -\lambda^{-1}I(\theta)^{-1}\nabla_\theta \Cat(\theta, \theta_p, f_\eta, \gamma)+ \theta,
\end{equation*}
where $I(\theta)$ denotes Fisher's information matrix.
\end{lemma}

A direct consequence of \Cref{lem:Cat_grad} is that \Cref{cor:main} applies when using the exact solver for the surrogate Catoni task. Since Fisher's information is positive, it follows that the update direction $\tilde{\theta}-\theta$ always diminishes the bound locally. We summarise these results in the following theorem.

\begin{theorem}
\label{corol:catoni}
The minimiser of Catoni's PAC-Bayes objective on an exponential family is a fixed point of \Cref{alg:general_surrog_PAC_Bayes} with approximation function 
\begin{equation*}
\Approx(\pi_\theta, \risk) := \arg\inf_{f\in\mathcal{F}} \pi_\theta[(\risk-f)^2],
\end{equation*}
and surrogate solver 
\begin{equation*}\Solve(\prior, \theta, f_\eta) := \theta_p - \lambda^{-1}\eta = \arg\inf_{\theta\in\Theta}\Cat(\theta, \prior, f_\eta, \gamma).
\end{equation*}

Moreover, for all $\theta$, 
\begin{equation*}\nabla \Cat \cdot (\Solve(\prior,\theta, \Approx(\theta, \risk)) - \theta) \leq 0.
\end{equation*}
\end{theorem}

As noted above, the solution of the surrogate task must belong to $\Theta$ to define a probability distribution. There is however no guarantee that such is the case for any approximated risk. For instance, if the risk is estimated close to a local maxima by a quadratic function, the resulting surrogate task might not have a minima, and hence the resulting $\theta(\eta)$ might fail to be a probability distribution, causing the algorithm to break. Another difficulty lies in solving the approximation task. Involving an integral of a function of the risk, the objective theoretically requires evaluations of the risk at all predictors.
We show in the next section how both these issues can be solved in practice.

\subsection{Framework implementation: \algo}
Following \Cref{corol:catoni}, we propose an algorithm, \algo (\url{https://github.com/APicardWeibel/surpbayes}), designed to efficiently find the minimiser of Catoni's bound on Exponential families. 

\subsubsection{Implementing the approximation}
\label{sec:Catoni_approx}
As the surrogate PAC-Bayes bound is solved using a closed form expression, the computational bottleneck of \Cref{alg:general_surrog_PAC_Bayes} is the approximation task of computing $\eta(\theta) = \arg\inf_{\R^d}\pi_\theta[(f_\eta -\risk - \pi_\theta[f_\eta-\risk])^2]$. Due to the form of $f_\eta$, this is formally a least square weighted linear approximation problem with infinite number of observations, whose solution can be explicitly written as
$\eta = \mathbb{V}_\pi[T]^{-1}\pi[\risk(T-\pi[T])].$ This solution can be approximated using a finite number of function evaluations $\risk(x_i)$, replacing the probability $\pi$ by an empirical counterpart $\pi_{\text{emp}}=\sum_{i=1}^N \omega_i \delta_{x_i}$.

Different choices of $(x_i, \omega_i)$ can be considered. A first approach consists in drawing i.i.d. samples from $\pi_\theta$ and considering uniform weights. This guarantees that the approximated objective is unbiased. A main shortcoming of this approach, however, is that it disregards all previous risk evaluations at each step. Corrections of the form $\frac{\text{d}\pi_\theta}{\text{d}\pi_{\tilde{\theta}}}$ can be used to salvage samples drawn from $\pi_{\tilde{\theta}}$, all the while guaranteeing unbiased approximated objective. This however can drastically increase the variance, and thus might not be practical.

We advocate a \emph{generation agnostic} approach for the weighing process, which treats all available risk evaluations in a like manner. We assume that $\X$ is a metric space. For all predictors $(x_i)_{i\in[1, N]}$ whose risk $\risk(x_i)$ is known, target weights $\tilde\omega_i$ are defined as the probability given to the Voronoi cell $\overline{x}_i$ by distribution $\pi_\theta$. This target weight can be approximated using Monte Carlo simulations and solving nearest neighbour in $(x_i)_{i\in[1, N]}$ tasks. The distance used for the Voronoi cell can depend on the distribution $\pi_\theta$ --(\emph{e.g.} Mahalanobis distance for Gaussian exponential families). This approach requires, if the empirical distribution $\sum \omega_i \delta_{x_i}$ is to form an adequate approximation of the distribution $\pi_\theta$, some queries from to $\pi_\theta$. The stack of function evaluation is hence appended at each approximation step by evaluating samples from $\pi_\theta$. As this weight computation can bring some overhead, it is only appropriate when risk queries are the main computational bottleneck.

\subsubsection{Boundary issues}
PAC-Bayes bounds typically hold for empirical risk functions satisfying moment bounds (with respect to the data generation mechanism) or boundedness conditions (the latter being usually required for Catoni's bound). 
Such assumptions might no longer be met for the approximated risks. A consequence is that the minimiser of the surrogate task might not exist. For instance, a local quadratic approximation of the score near a local maxima can induce a surrogate task whose minima is $-\inf$.

To ensure that for any score approximation $f_{\eta, C}$, the surrogate solver always define a probability distribution, two regularisation hyperparameters $\klmax$ and $\dampen$ are introduced. $\klmax\in\R_+\cup{+\infty}$ determines the maximum step size allowed between two successive posterior estimation, measured in Kullback--Leibler divergence. $\dampen\in ]0,1]$ acts as a dampening hyperparameter. The corrected update rule is changed to $\tilde{\theta}_c(\theta) = \tilde{\alpha}(\tilde{\theta}(\eta)-\theta) + \theta$ with $\tilde{\alpha}$ the highest $\alpha\leq \dampen$ such that $\KL(\tilde{\theta}_c, \theta) \leq \klmax$. Such $\tilde{\alpha}$ can be easily obtained through a Newton scheme or dichotomy, noticing that it is defined through $f(\tilde{\alpha}) = C$ for a non decreasing function $f$.

This modification does not impact the fixed point property of \Cref{corol:catoni}. Moreover, if the empirical risk $\risk$ belongs to $\mathcal{F}$, choosing $\klmax<\infty$, $\dampen = 1$ results in convergence in a finite number of steps (resp. exponential convergence for $\dampen<1$) (see \Cref{app:reg_convergence}).

\begin{remark}
While \algo is designed to optimize Catoni's PAC-Bayes bound \eqref{eq:catoni}, it can serve as a work engine for the minimisation of other PAC-Bayes bounds. For instance, Proposition 2.1 from \cite{germain-15a} implies that Maureer-Langford-Seeger's bound (MLS bound, \cite{maurer2004, langford2001bounds} can be rewritten as
\begin{equation*}
\text{PB}_{\text{MLS}} = \inf_{\lambda>0}\frac{
1 - \exp\left(
-\frac{\text{Obj}_{\text{Cat}, \lambda}}{\lambda n}
- \frac{\log(\xi(n)/\delta}{n} \right)
}{
1 - \exp\left(-1/(\lambda n)\right)
}.
\end{equation*}
As such, the minimisation of MLS bound could be performed by alternatively minimizing Catoni's objective at fixed temperature using \algo and solving on the temperature at fixed posterior. The generation agnostic weighing approach moreover implies that re-optimising Catoni's objective after a small change of temperature can be done with few new risk queries (see the strategy developped in \Cref{sec:Metalearn}). This strategy is further developed in \citet{picard2024note}.
\end{remark}

\section{Surrogate Catoni in a Meta-Learning framework}
\label{sec:Metalearn}
Both the Bayes and PAC-Bayes framework offer a natural connection with Meta-Learning, as both involve a natural inductive bias in the form of the prior. Previous work which studied Meta-Learning for PAC-Bayes include \citet{pentina2014pac,amit2018meta,rothfuss2023scalable,zakerinia2024more}. The aim of PAC-Bayes Meta-Learning is the construction, from a sample of independent train tasks, of a prior yielding optimal generalisation bounds on new unknown test tasks. Such optimisation of the prior brings two benefits: tighter generalisation bounds (smaller penalisation); and simplified PAC-Bayes learning task (better initial guess). For PAC-Bayes meta learning, a natural training objective can be derived from the minimised PAC-Bayes bounds obtained for each task. This defines the following meta training objective, analogue to an empirical risk at the meta level:

\begin{equation}
\label{eq:meta_objective}
M(\prior) = \sum_{i} \inf_{\pi\in \Pi} \PB(\hat{\pi}_i, \risk_i, \prior, \eta_i),
\end{equation}
where $\hat{\pi}_i$ denotes the task posterior and is a function of $R_i$, $\prior$ and $\eta_i$. The objective defined in \Cref{eq:meta_objective} departs from previous formulations which typically involve a further penalisation term at the meta level. We advance two justifications for this simplification. First, the extra penalisation term involves divergence terms between a meta prior and meta posterior (both distributions on probability distributions) which in practice make the bound vacuous and thus of limited practical interest. Second, PAC-Bayes theory already offers guarantees on the generalisation performances of each test task, limiting the need to assess the generalisation performance at the meta level. Arguably, the task specific bound provided by using PAC-Bayes as inner algorithm is more informative than the "mean" task bound offered by a meta PAC-Bayes algorithm (when PAC-Bayes learning is used both as inner algorithm and meta training algorithm).

We consider that assumptions \emph{($A_1$)} to \emph{($A_5$)} hold, and also these further mild assumptions: the prior is looked for in $\Pi$, i.e $\pi= \pi_{\theta_p}$; the PAC-Bayes bound $\PB$ is differentiable w.r.t. $\theta_p$. Then, noting $\hat{\theta}_i$ the posterior parameter for each task, a simplification of the meta gradient occurs:
\begin{equation}
\label{eq:meta_gradient}
\nabla M(\theta_{p}) = \partial_{\theta_p} \PB(\hat{\theta}_i(\theta_p), R_i, \theta_p, \eta_i)= \partial_3 \PB(\hat{\theta}_i, R_i, \theta_p, \eta_i).
\end{equation}
Remarkably, the knowledge of the derivative of $\hat{\theta}_i$ with respect to $\theta_p$ is not required to compute the meta gradient. This is due to $\partial_1 \PB$ being $0$ when evaluated for the prior posterior. We stress that such a simplification is specific to our meta-learning objective. It does not occur in meta-learning strategies such as MAML \citep{finn17a}, where the performance of each task is assessed on a test set. In the context of PAC-Bayes, such reliance on test sets can be optimistically replaced by the PAC-Bayes bounds, which give test guarantees with high probability. It is unclear whether such a simplification occurs in previous PAC-Bayes Meta Learning objectives from the literature, as these involve distributions on priors rather than a single prior.

A key consequence is that training the meta learning algorithm is as hard as cycling all the Bayesian optimisation tasks. In a nutshell, meta learning is as hard as re optimising the bound for a new prior.

\algo brings two main benefits when used in conjunction with meta-learning. First, by improving the optimisation efficiency for a given prior, \algo speeds up the meta-learning procedure. Second, the "generation agnostic" weighing approach implies that risk revaluations from previous optimisation procedures can be reused. As a consequence, re optimisation of a PAC-Bayes bound for a new prior can conceivably be performed with few risk queries, bringing an additional speed-up. Moreover, the setting considered for \algo enjoys an analytical expression for meta-gradients, $
\nabla M(\theta_p) = \sum_i\lambda_i (\nabla g(\hat\theta_i) - \nabla g(\theta_{p}))$ which can be efficiently evaluated.

\section{Experiments}
\label{sec:exp}

\begin{figure}
    \centering
    \subfloat[Optim. perf.]{
\includegraphics[width=.505\textwidth]{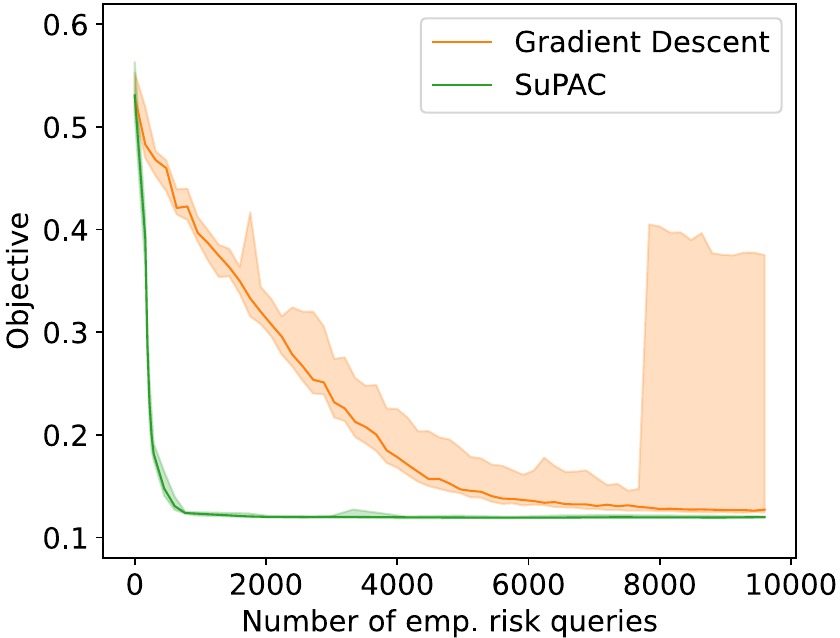}
\label{subfig:optim_perf}
    }
    \subfloat[Meta Learning]{
\includegraphics[width=.443\textwidth]{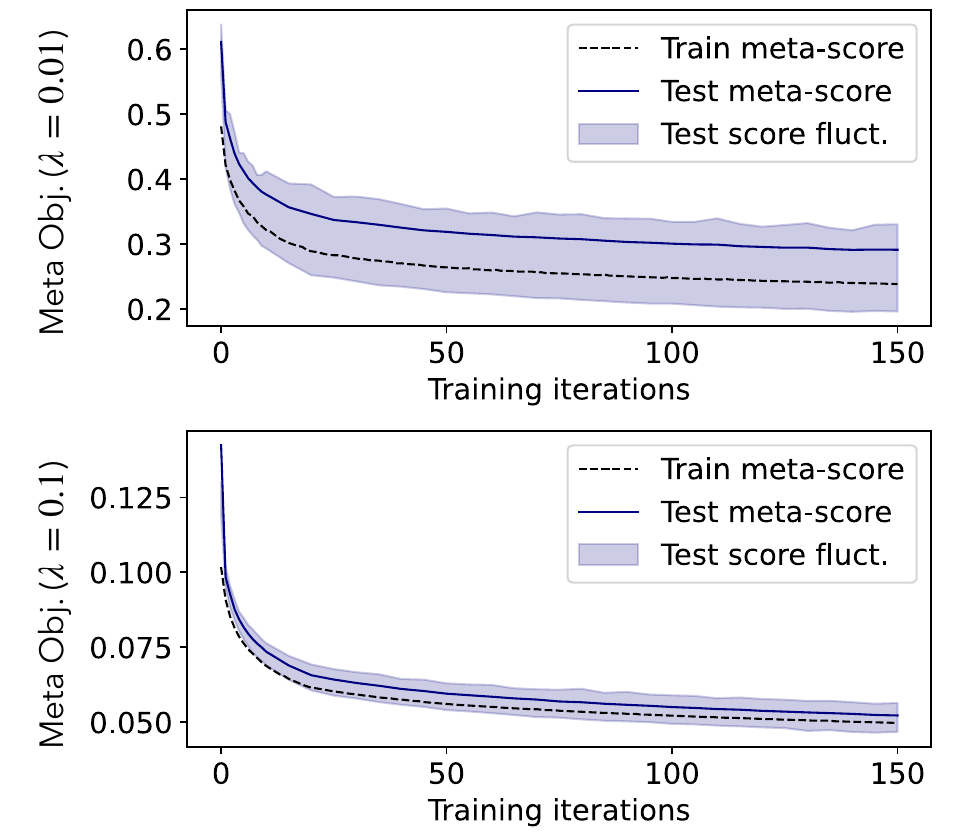}
\label{subfig:meta}
    }
\label{fig:enter-label}
\caption{Experiments results. \Cref{subfig:optim_perf} compares the optimisation performance of our algorithm \algo with gradient descent approaches on an biochemical calibration task. Optimisation procedures were repeated 20 times; median performance and quantiles 0.2 and 0.8 are represented. \Cref{subfig:meta} investigates train and test performance of the meta-learning approach of \Cref{sec:Metalearn}. Mean test performance, as well as quantiles 0.2 and 0.8 for the sequence of built prior is assessed on 40 tasks and compared to the train performance. \algo reduced the PAC-Bayes objective to $0.121\pm 0.004$ (avg. risk of posterior of $0.102\pm 0.003$).}
\end{figure}

\algo was assessed on the learning task described by \cite{PicardWeibel2024}. A PAC-Bayes bound is minimised on Gaussian distributions with block diagonal covariance in order to calibrate 30 parameters of a biological inspired numerical model describing anaerobic digestion processes, ADM1 \citep{Batstone2002}. This model relies on solving a stiff ODE to predict the evolution of the states, and is therefore quite computationally intensive (about 3 seconds per model query in our experiments).

We compared \algo to standard GD on a synthetic dataset from \cite{PicardWeibel2024}, using the same family of distributions and risk function. For \algo, 160 risk queries where performed for the initial step, and 32 for all further step. A maximal budget of 9600 empirical risk queries was fixed; hyperparameters for the GD were selected after evaluating a grid on the first 1600 queries. Mean risks were assessed at test time by resampling new predictors from the posterior. The PAC-Bayes temperature was set to $0.002$. Training procedures were repeated 20 times.

The performance of the sequence of posteriors were compared by aligning the number of empirical risk queries. Indeed, the main motivation of \algo is the setting when querying the empirical risk is computationally expensive, and can be assumed to be the computational bottleneck. This is indeed the case for the anaerobic digestion example considered here. At equal number of risk queries, \algo required an extra 3.5\% processing time compared to gradient descent, mainly caused by the weighing process.

\algo proved significantly more efficient at minimising the bound than GD (see \Cref{subfig:optim_perf}). The average performance of our algorithm proved better after 1800 queries than the best performance obtained after the full 9600 queries for GD. The experiments also indicate that our procedure offers much higher stability compared to GD, both during training and between the training duplicates. This could be attributed to the "generation agnostic" weighing approach, which relies on all previous risk evaluations at each step and is thus more stable. On the other hand, the noisy gradients estimates have some probability of leading to problematic steps during GD, leading to sharp increase in the objective. In our experiments, 4 out of 20 GD procedures thus led to a worse performance than the one obtained by a single optimisation step of \algo. The posterior distributions constructed through \algo obtained an average empirical risk of $0.102 \pm 0.003$, similar to the $0.101$ value reported in \cite{PicardWeibel2024}. The resulting PAC-Bayes bound proved also similar ($0.121\pm 0.004$ vs. $0.122$). Thus \algo constructed as good a posterior as \cite{PicardWeibel2024}, but twenty times faster.

Further assessments of \algo's performance for other hyperparameters values and comparison to Nesterov accelerated GD were also conducted. \algo proved to have a stable behaviour for a wide range of hyperparameters value ($0.25 \leq \dampen \leq 0.75$, $0.5 \leq \klmax\leq 2$), with instabilities starting to appear for $\klmax > 5$, and speed decrease for $\klmax<0.1$. Nesterov acceleration, requiring some iterations to build up momentum, proved unable to compete with \algo's almost instantaneous optimisation. Results for these experiments can be found in \Cref{app:implement}.

Preliminary experiments were also performed for the meta-learning objective described in \Cref{sec:Metalearn}. To facilitate the evaluation of the learnt meta priors, wholly synthetic risk functions were used in this case, and PAC-Bayes objective minimised on Gaussian distributions. The risk functions considered were bounded, smooth functions of $\mathbb{R}^8$, achieving their global minima at $x_0\sim \mathbb{N}(\tilde{x}_0, \Sigma_0)$. $\tilde{x}_0$ was chosen so that $\lVert\tilde{x}_0\rVert = 2$, and $\Sigma_0$ such that only two of its eigenvalues are higher than $0.05^2$ (drawn at random between $\exp(-1)$ and $\exp(1)$).

Such choices ensure that the original prior distribution, $\mathcal{N}(0, I_k)$, can be improved upon both by shifting its mass centre and adjusting its covariance.
The performance of the meta-learning algorithm was assessed for two temperatures, $\lambda=0.1$ and $\lambda=0.01$. Meta training was performed using stochastic gradient descent. The sequence of prior thus constructed was evaluated on a further 40 test tasks, each time restarting the optimisation procedure from scratch, and evaluating the final score on $10^4$ draws from the posterior.

The meta-learning algorithm was able to satisfactorily reduce the objective, from an initial average generalisation bound of 0.61 (resp. 0.14) to 0.24 (resp. 0.050) after 150 gradient steps for $\lambda=0.1$ (resp. $\lambda=0.01$). Most of the meta-objective reduction takes place during the early phase of training, with the first 15 steps amounting to more than 80 \% of the objective decrease. For both temperatures tested, the average performance on the test tasks followed the objective decrease throughout training, even though the number of queries per optimisation was minimal after the first meta step (less than 40), supporting both our meta-learning objective and the use of \algo.

Full implementation details on the experiments can be found in \Cref{app:implement} and in the source code (\href{https://github.com/APicardWeibel/surpbayes}{https://github.com/APicardWeibel/surpbayes}).

\section{Discussion}
\label{sec:Disc}

The present work shows that it is possible to locally decouple the complexity related to querying the empirical risk and the minimisation of a PAC-Bayes bound. A main motivation for such decoupling is that the approximated risk function defines a non linear surrogate objective which might be valid (\emph{i.e.} close to the original objective) for a wider range of probabilities than the linear approximations offered by the gradients. As a consequence, the surrogate bound solution can be reasonably allowed to be much further away from the current posterior estimation than is the case for GD. A key implementation difficulty remains picking the range of validity, \emph{i.e.} how far away from the current posterior the surrogate solver can be allowed to choose a distribution. Such a choice, formalised in the selection of an adequate surrogate solving algorithm, is analogue to the choice of a step size in gradient descent procedures, and balances the stability and speed of the procedure. Automating the selection of the surrogate validity range offers an exciting prospect for the framework.

The Voronoi cell weighing approach used to solve the approximation problem is equivalent to replacing the empirical risk function by a 1-nearest neighbour trained predictor, and approximating this predictor. Variants following this two step approximation approach could be worth investigating. Notably, an interesting perspective would be to approximate the empirical risk through Gaussian processes, taking inspiration from Gaussian Optimisation. This would notably track the uncertainty on the approximate risk on extrapolated values, which could drive the choice of new predictors to evaluate and improve on the current random draws.

A key restriction of the present work is that our surrogate PAC-Bayes framework is only practicable when the dimension of the predictor space and of the probability family are small (\emph{i.e.} less than a few hundreds). This is due to two factors; first of all, the larger the dimension of the probability family, the larger becomes the approximation space, and hence the more empirical risk evaluations are required. Notably, at least $d+1$ evaluations of the empirical risk are required for probability families of dimension $d$. The second factor is that the "generation agnostic" weighing approach described in \Cref{sec:Catoni_approx} is unlikely to give adequate performances if $\mathcal{H}$ is high dimensional. This effectively rules out deep learning settings, which have been recently the main focus of the PAC-Bayes community. Still, we believe that PAC-Bayes learning offers meaningful prospects for a wide range of physics, biology or medical inspired problems which involve few parameters and expensive model computations, and therefore can be efficiently trained using our framework. Concrete fields of application of \algo include, but are not limited to, fluid dynamics simulations with dimension reduction \citep{Callaham2021}, metabolic models for microbial communities \citep{Cerk2024} and greenhouse gas emission inverse problems \citep{Nalini2022}.
We remark that, as of now, PAC-Bayes has not been much used outside of the learning community. While this can be vastly attributed to a lack of awareness of PAC-Bayes theory outside of the learning community, the use of PAC-Bayes was also hampered by the fact that previous PAC-Bayes algorithm required a prohibitive number of simulations and hence computation time. We believe \algo is a game changer in that respect, due to its focus on limiting the number of risk queries, and readily usable implementation, and we hope that this can be leveraged in different disciplines.

\paragraph{Conclusion.} We introduced a generic framework for minimising PAC-Bayes bounds designed to tackle computationally intensive empirical risks for low to moderate dimensional problems such as naturally arise in physical models. We established that our optimisation strategy was theoretically well supported. We instantiated this framework for the optimisation of bounds on exponential family, and considered how this implementation could interact with meta-learning. Preliminary experiments showed that our framework could significantly reduce the number of empirical risks queries when calibrating a biochemical model, thus opening exciting new fields of applications for PAC-Bayes.

\acksection{We warmly thank reviewers and the Area Chair who provided insigthful comments and suggestions which greatly helped us improve our manuscript. A.P. acknowledges support by ANRT CIFRE grant 2021/1894. B.G. acknowledges partial support by the U.S. Army Research Laboratory and the U.S. Army Research Office, and by the U.K. Ministry of Defence and the U.K. Engineering and Physical Sciences Research Council (EPSRC) under grant number EP/R013616/1. B.G. acknowledges partial support from the French National Agency for Research, through grants ANR-18-CE40-0016-01 and ANR-18- CE23-0015-02, and through the programme “France 2030” and PEPR IA on grant SHARP ANR-23-PEIA-0008.}

\bibliographystyle{plainnat}
\bibliography{bib}

%%%%%%%%%%%%%%%%%%%%%%%%%%%%%%%%%%%%%%%%%%%%%%%%%%%%%%%%%%%%

\clearpage
\appendix
\appendix

\section{Technical proofs}
\label{app:proof}

\subsection{Proof of \Cref{cor:main}}
\label{app:cor_proof}
As assumptions ($A_1$) to $(A_5)$ hold, \Cref{thm:flat_approx} can be used. It implies that replacing $R$ by $f^{R,\theta}$ does not change the gradient of $\text{PB}$. Hence, starting from $\theta=\theta^*$, since $\partial_1 \PB(\theta^*, R, \prior, \gamma) = \partial_1 \PB(\theta^*, f^{R, \theta^*}, \prior, \gamma) = 0$, the update step in the solving strategy satisfies $\text{Updt}(\theta^*)=\theta^* - M(\pi, \theta, f, \gamma) \times 0 = \theta^*$. Hence, by recursion, it follows that $\Solve(\prior, \pi_{\theta^*}, f^{R,\theta^*}) = \pi_{\theta^*}$. Since we assume that $\Approx(\pi_\theta, R) = f^{R,\theta}$, this implies that $\pi^{\theta^*}$ is a fixed step of $\pi \rightarrow \Solve(\prior, \pi, \Approx(\pi, R)))$, and hence that the posterior is a fixed point of \framework for the specified $\Approx$ and $\Solve$ strategies, concluding the proof.

\subsection{Proof of \Cref{lem:Cat_grad}}
\label{app:cat_grad_proof}
We consider the broader problem where the prior $\prior$ might not belong to the exponential family, but any probability satisfying the following assumptions:

\begin{hyp}
\label{Hyp:further}
\begin{enumerate}[label=(\subscript{A}{{\arabic*}})]
\setcounter{enumi}{5}
\item $\prior$ is absolutely continuous with respect to $\pi_{\text{ref}}$;
\item $\forall\theta\in\Theta$, $h:= \log\left(\frac{\text{d}\prior}{\text{d}\pi_{\text{ref}}}\right)\in\text{L}^2(\pi_\theta)$.
	\end{enumerate}
\end{hyp}

Note that when $\prior\in \Pi$, one can use $\pi_{\text{ref}}=\pi_p$ for which assumptions \emph{($A_6$)} and \emph{($A_7$)} are automatically fulfilled.
The generalisation of the approximation space becomes 
\begin{equation*}\mathcal{F} = \{f_{\eta, C}:= \theta\cdot T + C + \lambda h\},
\end{equation*}
which fits into the framework described in \Cref{thm:main}. For any $f_{\eta}\in\mathcal{F}$, the solver of Catoni's bound on all distributions is given by $\tilde{\theta} = -\lambda^{-1}\eta$, provided this defines a probability distribution (else Catoni's bound does not reach its minima on $\Pi$ or $\mathcal{P}$). Note that the choice of $\tilde{\theta}$ is coherent with the formula given in \Cref{lem:Cat_grad} when the prior belongs to $\Pi$, since in that case $h = \theta_p \cdot T$, leading to a change of coordinate in the definition of $\mathcal{F}$.

Under the assumptions, Catoni's bound is differentiable and its gradient with respect to $\theta$ can be computed under the integral. Thus, for score $f_\eta$,
\begin{align*}
\nabla \Cat &= \pi_\theta[f_\eta (T-\nabla g(\theta))] + \lambda \pi_\theta[(\theta\cdot T - g(\theta) -h)(T-\nabla g(\theta))] \\
&= \pi_\theta[(f_\eta+ \lambda \theta\cdot T - g(\theta)- \lambda h)(T-\nabla g)]\\
&= \pi_\theta[(f_\eta+ \lambda \theta\cdot T - \lambda h)(T-\pi_\theta[T])]\\
&= \pi_\theta[(\eta\cdot T + C)(T-\pi_\theta[T])] + \lambda \mathbb{V}_{\pi_\theta}[T]\theta\\
&= \mathbb{V}_{\pi_\theta}[T](\eta + \lambda \theta)
\end{align*}
where we use the well known identity $\pi_\theta[T] = \nabla g$ (see \cite{Brown1986}). For exponential families, the variance $\mathbb{V}_{\pi_\theta}[T]$ coincides with Fisher's information, and hence the previous equality reads $\nabla\Cat = \lambda I(\theta)(\theta - \tilde{\theta}(\eta))$, which implies \Cref{lem:Cat_grad}.

\subsection{Probability families with constant approximation space}
\label{app:constant_F}
\Cref{thm:main} considers projections of the risk on a local vector space of functions $\mathcal{F}_\theta$. A special case of interest concerns families of distributions such that the approximation set is constant. Exponential families offer such a characteristic. We show here that exponential families (and its restrictions) are the only smoothly parameterised distributions with this characteristic:

\begin{theorem}
\label{thm:flat_approx}
For family of distributions satisfying the first three hypotheses of \Cref{Hyp:std} such that, moreover:
\begin{itemize}
	\item $\Theta$ is a connected,
	\item $\theta\rightarrow\ell(\theta,x)$ is twice continuously differentiable for all $x$.
	\end{itemize}

If there exists a vector space of finite dimension $\mathcal{F}$ such that $\mathcal{F}_\theta\subset\mathcal{F}$ for all $\theta\in\Theta$, then there exists an exponential family $\Pi_{T}$ defined on $\widetilde{\Theta}$ and a connected, open set $\Theta_{\Pi}$ such that $\Pi= \{\pi_\theta \mid\theta\in\Theta_{\Pi}\}$.
\end{theorem}

\begin{proof}
For $\mathcal{F}$ of dimension $\tilde{d}+1$, choose $T_1,~\dots,~ T_{\tilde{d}},~T_{\tilde{d}+1}=1$ a basis of $\mathcal{F}$. Then, for all $\theta$, there exists a unique matrix $A(\theta)\in \R^{d, \tilde{d}}$, and a unique vector $c\in \R^{\tilde{d}, 1}$ such that

\begin{align*}
\partial_\theta \ell = \begin{pmatrix}A(\theta) & c(\theta) \end{pmatrix} \begin{pmatrix}T_1\\\cdots\\T_{\tilde{d}+1}\end{pmatrix}
\end{align*}

Assume that $A(\theta)$ and $c(\theta)$ are differentiable (this is proved afterwards).
Since $\ell$ is twice continuously differentiable, it follows $\partial_{\theta_i}\partial_{\theta_j}\ell = \partial_{\theta_j}\partial_{\theta_i}\ell$, and therefore that
$\partial_{\theta_{i}}A_{j,k} = \partial_{\theta_{j}}A_{i,k}$ and that $\partial_{\theta_j} c_i = \partial_{\theta_i} c_j$. This, in conjunction with the hypothesis that $\Theta$ is connected, implies that $A(\theta)$ is a gradient of some $\beta:\R^{d}\mapsto\R^{\tilde{d}}$ while $c$ is the gradient of some $-g:\R^{d}\mapsto\R$ (see \cite{Lang1999}). Hence, $\ell(\theta) = \beta(\theta) \cdot T(x) - g(\theta) + h$ for $h$ a solution of $\partial_\theta h = 0$. Since $\Theta$ is connected, this implies that $h$ can not be a function of $\theta$. Hence $\Pi$ is the restriction of an exponential family on $\Theta$.

It remains to show that $A(\theta)$ and $c(\theta)$ are differentiable.
First of all, we remark that for all finite collection of linearly independent real valued functions $(f_1, \dots, f_{n})$, there exists $d$ points $(x_{1}, \dots, x_{n}$ such that $(f_i(x_j))_{i,j\leq n}$ is inversible. Indeed, this result holds for a single function, since $f_1$ must be non zero. Then if the result holds for $x_1, \dots, x_k$, \emph{i.e.} $D = \text{det}( (f_i(x_j))_{i,j\leq k}) \neq 0$ then consider the matrix $m(z) = (f_i(\tilde{x}_j)_{i,j\leq k+1}$ with $\tilde{x}_j= x_j$ if $j\leq k$, $\tilde{x}_{k+1} =z$. Then the determinant of matrix $m$ is $D f_{k+1}(z) + \sum_{i\leq k} C_i f_i(z)$. Since $f_1, \dots, f_{k+1}$ are linearly independent and since $D$ is not zero, there must exist $z$ such that $\text{det}(m(z)) \neq 0$, which we can pick as $x_{k+1}$. This proves the result by recursion.

Since $T_1, \dots T_{\tilde{d}+1}$ are linearly independent, we can therefore pick such $x_1, \dots, x_{\tilde{d}+1}$. By definition of $A(\theta)$ and $c(\theta)$, it follows that for all $\theta$, 
\begin{align*}
\begin{pmatrix}
A(\theta)& c(\theta)
\end{pmatrix}
=
\begin{pmatrix}
\partial_{\theta_1} \ell(\theta, x_1) & \dots &\partial_{\theta_1} \ell(\theta, x_{\tilde{d}+1})\\\vdots & &\vdots\\
\\
\partial_{\theta_k}\ell(\theta, x_1) & \dots &\partial_{\theta_k} \ell(\theta, x_{\tilde{d}+1})
\end{pmatrix}
\begin{pmatrix}
T_1(x_1) &\dots & T_1(x_{\tilde{d}+1})\\
\vdots & &\vdots\\
T_{\tilde{d}+1}(x_1) & \dots & T_{\tilde{d}+1}(x_{\tilde{d}+1}) 
\end{pmatrix}^{-1}
\end{align*}

This implies that $A$ and $c$ are linear combinations of the differentiable functions $(\partial_\ell(\cdot, x_{i}))_{i\in[1,\tilde{d}+1]}$, and hence that they are differentiable.

\end{proof}

\subsection{Regularisation and convergence for Catoni's bound}
\label{app:reg_convergence}
If $\risk = f_\eta\in\mathcal{F}$, the uncorrected step direction results in one step convergence, implying that the update direction at $\theta$ is $\hat{\theta} - \theta$. This implies that all successive estimation $\theta_i$ belongs to the segment $[\theta_0, \hat{\theta}]$. Note $\Delta \theta = \hat{\theta} - \theta_0$. Since the normalisation function $g$ is strictly convex, it follows that the function $t\rightarrow \Delta \theta \cdot \nabla g(\theta_0 + t \Delta\theta)$ is non decreasing, and hence, for all $t$,
\begin{equation*}\Delta \theta \cdot\nabla g(\theta_0) \leq \Delta \theta \cdot \nabla g(\theta_0 + t \Delta\theta)\leq \Delta \theta \cdot \nabla g(\hat{\theta}).
\end{equation*}

Using the convexity of $g$, this implies that for $t_1 < t_2$, $g(\theta_0 + t_1\Delta_\theta) - g(\theta_0 + t_2\Delta_\theta)\leq (t_1-t_2) \Delta_\theta\cdot \nabla g(\theta_0)$ while $(t_2 - t_1)\Delta_\theta \cdot \nabla g(\theta+ t_2\Delta\theta)\leq (t_2-t_1) \Delta_\theta\cdot\nabla g(\hat{\theta})$.

It follows that for all $t_1< t_2$,
\begin{equation*}
\KL(\theta_0 + t_2 \Delta\theta, \theta_0 + t_1\Delta\theta) \leq (t_2 - t_1) \Delta\theta\cdot(\nabla g(\hat{\theta}) - \nabla g(\theta_0)).
\end{equation*}
This implies that for $\theta_i= \theta_0 +t_i \Delta\theta$, $\theta_{i+1} =\theta_0 + t_{i+1}\Delta\theta$, if the condition $\KL(\theta_{i+1}, \theta_i) \leq \klmax$ is active, then $t_{i+1} - t_{i} \geq \frac{\klmax}{\Delta\theta\cdot(\nabla g(\hat{\theta}) - \nabla g(\theta_0))}$. Since $t_{i+1} - t_i\geq 0$ and for all $i$, $t_i\leq 1$, this implies that the condition is active a finite number of time at most. In the case of $\dampen = 1$, this implies convergence in a finite number of steps. For $0\leq\dampen< 1$, this implies that after some $K$, $t_{i+K} = (1-\dampen)^{i} (1 - t_K)$, and hence exponential convergence of $(\theta_i)$ to $\hat{\theta}$.

\section{Implementation details}
\label{app:implement}
The code described in this section can be found in the publication repo: \url{https://github.com/APicardWeibel/surpbayes}.
\subsection{Further notes on \algo}
\algo can be summarised in the following pseudo-code:
%\begin{minipage}{.6\linewidth}
\begin{algorithm}[H]
\caption{Surrogate Catoni solver for exponential families (\algo)}\label{alg:catoni_solver}
\begin{algorithmic}
\Require $\lambda > 0$, $\theta_0\in\Theta$, $\theta_p\in\Theta$, $\risk\in\mathcal{M}(\X)$, $\text{Ev} = (x_i, \risk(x_i))_{i=1}^n$, $0<\dampen\leq 1$, $0<\klmax$
\State $\theta \gets \theta_0$
\While{not converged}
	\State Draw i.i.d. $x_{n + 1}, \dots, x_{n+k} \sim \pi_{\theta}$
    \State $\text{Ev}, n \gets \text{Ev}\cup ((x_{n+1}, R(x_{n+1})), \dots, (x_{n+k}, R(x_{n+k}))), n+k $
    \State $\omega_i \gets \pi[\overline{x}_i]$ \Comment{Solving nearest neighbour problems}
    \State $\eta^*, C  = \arg\inf_{\eta,C}\sum_{i\leq n} \omega_i (T(x_i) - \risk(x_i) -C)^2$
    \State $\delta\theta = \theta_0 - \lambda^{-1}\eta^* - \theta$ 

    \State $\tilde\alpha\gets \sup\{\alpha \mid\alpha<\dampen, \KL(\theta+ \alpha\delta\theta, \theta)\leq\klmax\}$
    \State $\theta\gets\theta + \tilde{\alpha}\delta\theta$
\EndWhile
\end{algorithmic}
\end{algorithm}
%\end{minipage}

Our implementation is based on the pre-existing code source provided by \cite{PicardWeibel2024}. Part of the original code was reworked to fit our new setting. New classes for exponential families of distributions were introduced, and implementation of the Gaussian family classes modified accordingly. A modular and generic solver class for the minimisation of Catoni's PAC-Bayes bound on exponential families was introduced, as well as more specific implementations for probability families outputting Gaussian distributions, using the Mahalanobis distance when approximating the weights. These solvers rely on closed form expressions for the Kullback--Leibler divergence and its derivative, inferred from the normalisation function and its derivatives.

The default weighing approach for the score approximation uses exact 1-NN for a user specified number of samples ("n\_estim\_weights" argument), performed using Faiss library \citep{faiss-lib}. Another weight approximation method, relying on approximate k-NN solving, is also provided.

The corrected update rule parameter $\tilde{\alpha}$ is estimated by dichotomy, using the fact that for all $\theta$, $\delta\theta$, the function $\alpha\rightarrow \KL(\theta+\alpha\delta\theta, \theta)$ is not decreasing. The resulting $\tilde{\alpha}$ is guaranteed to result in a Kullback--Leibler step of less than $\klmax$.

\label{app:algo}
\begin{figure}
  \centering
  \includegraphics[width=0.7\textwidth]{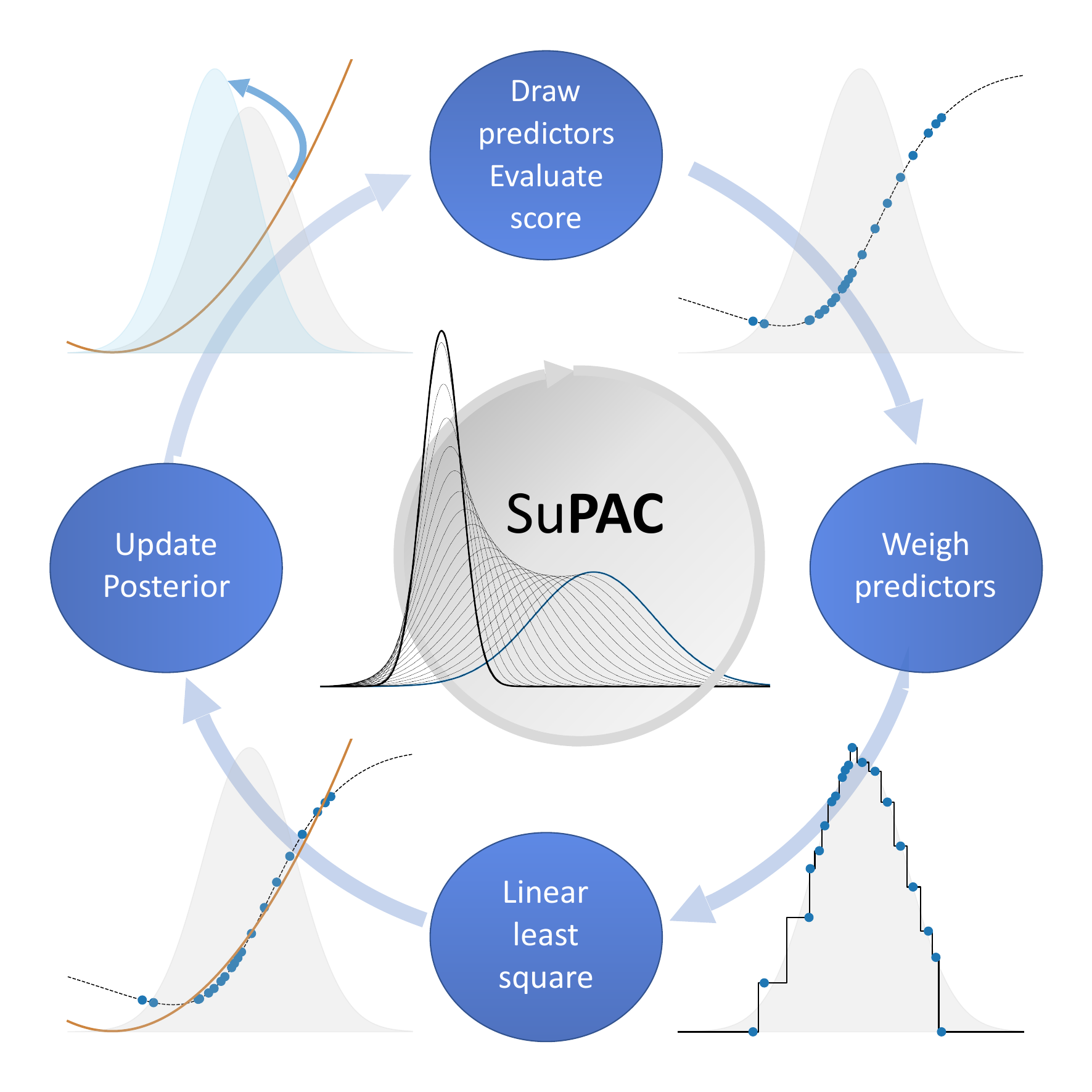}
  \caption{Overview of \algo. At each step, some new predictors are drawn from the current posterior approximation and evaluated (top right figure). All evaluated predictors are then weighted according to the weight of their Voronoi cell (bottom right figure). These weighted evaluations are used to construct an optimal approximation of the score through a linear least square task (bottom left figure). The approximated score is used to update the posterior using a closed form expression (top left figure). This procedure is looped until convergence (center).}
 \end{figure}

 \subsection{Experiments}
\subsubsection{Catoni's bound minimisation}
The implementation of ADM1 from \cite{PicardWeibel2024} was used to perform the experiments, and slightly modified to benefit from just-in-time compilation. The dataset used was the training part of dataset "LF". The probability family (Gaussian with block diagonal covariance with fixed blocks) and prior distribution considered in the original paper was used.
For \algo, the regularisation hyperparameters were set to $\klmax=1$ and $\dampen=0.5$, while the number of samples generated to evaluate the weights was set to 40 000. The optimisation algorithm was trained on 296 steps; for the initial step, 160 risk queries were performed, while for all the remaining steps, 32 risk queries were performed. This larger number of queries for the initial step is due to the necessity of having a least more evaluations than the dimension of the family of probability.

\begin{figure}
\centering
\includegraphics[width=.7\textwidth]{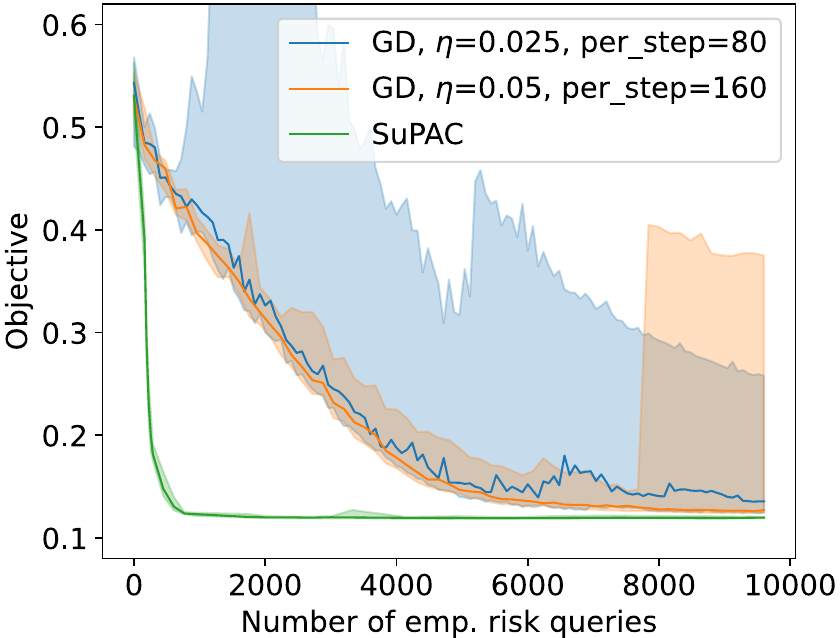}
\label{fig:full_comp_optim}
\caption{Comparison of the optimisation procedures as performed by \algo and gradient descent (GD) for the two selected sets of hyperparameters.  Each optimisation procedure was repeated 20 times; the median performance and 0.2 and 0.8 quantiles are represented. \algo was performed with hyperparameters $\dampen = 0.5$ and $\klmax=1$.}
\end{figure}

Hyperparameters for GD were selected after assessing the grid $\text{per\_step} \in \{80, 160\}$, $\text{step\_size} \times \{0.025, 0.05, 0.07\}$ on a preliminary 1600 score queries budget, with 20 repeats. The larger step size 0.07 was rejected due to its erratic behaviour between repeats, obtaining both optimal and worse GD performance. This erratic behaviour was also observed for step size 0.05 when estimating gradients from 80 risk queries. On the other hand, for per\_step set to 160, the step size of $0.025$ clearly under-performed compared to the step size of $0.05$, although slightly more stable. This led to the selection of the two sets of hyperparameters, (per\_step=80, step\_size=0.025) and (per\_step=160, step\_size=0.05), which had similar performances. Both were assessed, and the set of hyperparameters obtaining the lowest score, (per\_step=160, step\_size=0.05), was kept for comparison (see \Cref{fig:full_comp_optim}).

\begin{figure}
\centering
\subfloat[$\eta$ = 0.025]{
\includegraphics[width=.31\textwidth]{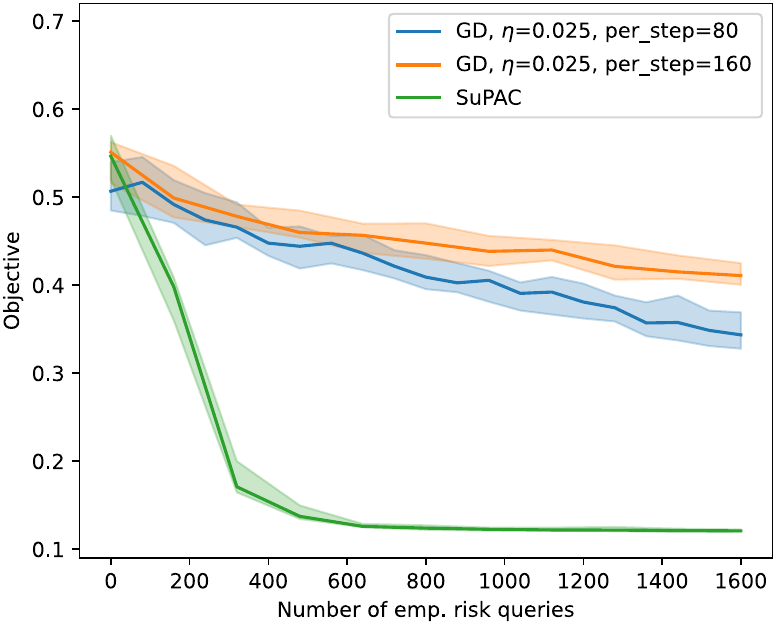}
\label{subfig:prelim_025}
}
\subfloat[$\eta$= 0.05]{
\includegraphics[width=.31\textwidth]{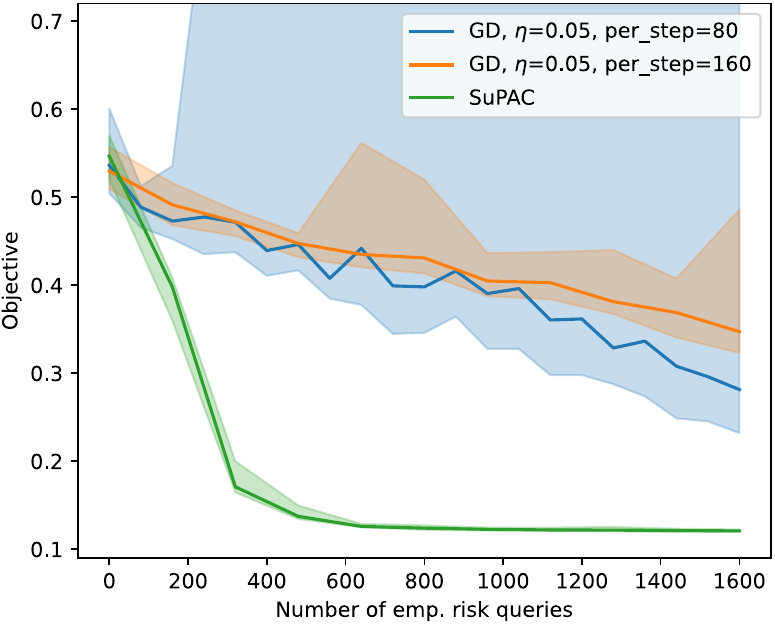}
\label{subfig:prelim_05}
}
\subfloat[$\eta$ = 0.07]{
\includegraphics[width=.31\textwidth]{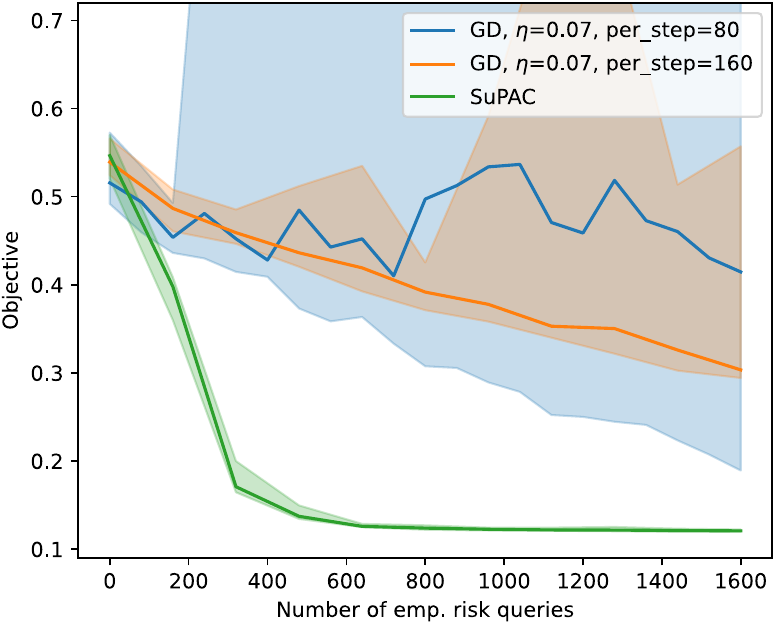}
\label{subfig:prelim_07}
}
\caption{Preliminary GD optimisation procedures for different choices of hyperparameters. The evaluations of each optimisation procedure was repeated 20 times; the median performance and 0.2 and 0.8 quantiles are represented. The performance of \algo is given for comparison.}
\end{figure}

\begin{figure}
\centering
\includegraphics[width=\textwidth]{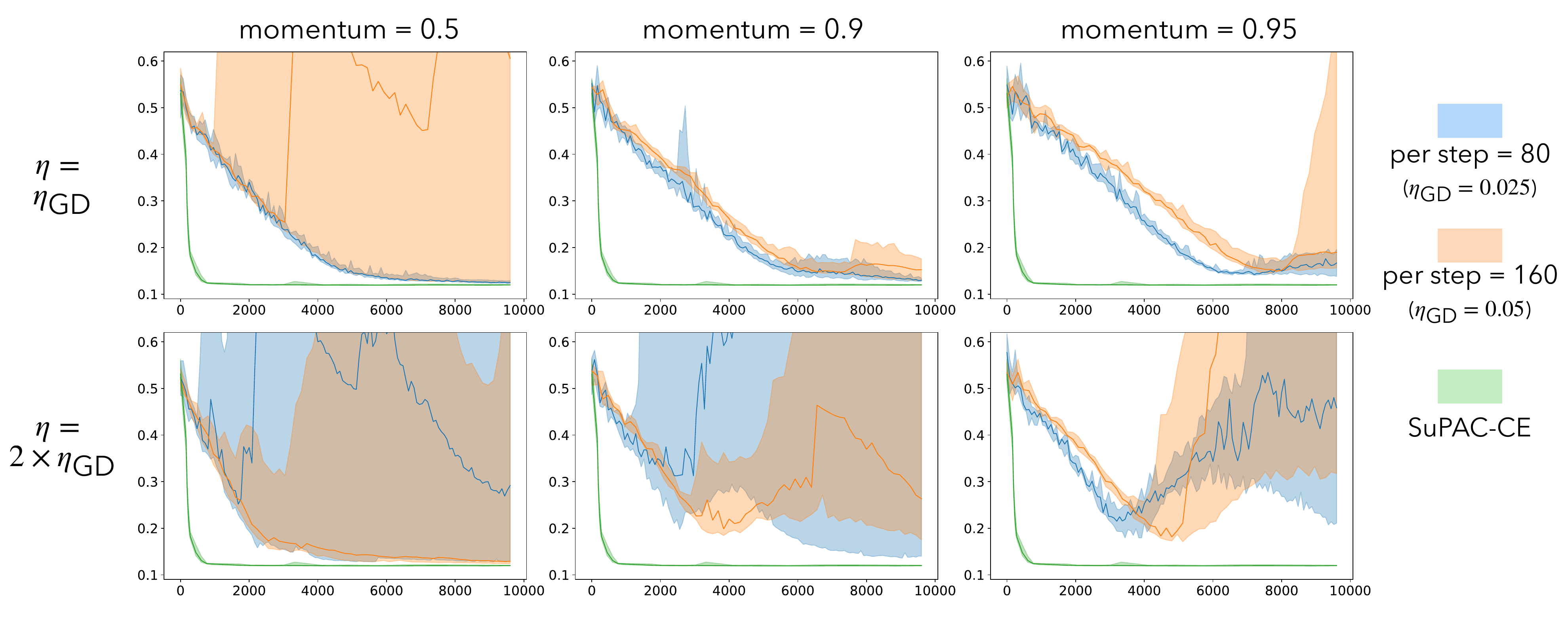}
\label{fig:nesterov}
\caption{Comparison of the optimisation procedures as performed by \algo and Nesterov accelerated gradient descent (x axis: number of empirical risk queries). Each optimisation procedure was repeated 8 times; the median performance and 0.2 and 0.8 quantiles are represented. \algo was performed with hyperparameters $\dampen = 0.5$ and $\klmax=1$. Momentum of $0.5$, $0.9$ and $0.95$ were assessed for Nesterov gradient descent. Both the original step size ($\eta$) parameter as well as twice the step size parameter for gradient descent comparisons were investigated. At twice the step size, all momentum accelerated procedures proved unstable. At the original step size, the momentum tended to increase the stability of the procedure at the cost of speed. All Nesterov accelerated gradient descent procedures assessed were slower than \algo}
\end{figure}

\begin{figure}
\centering
\includegraphics[width=.95\textwidth]{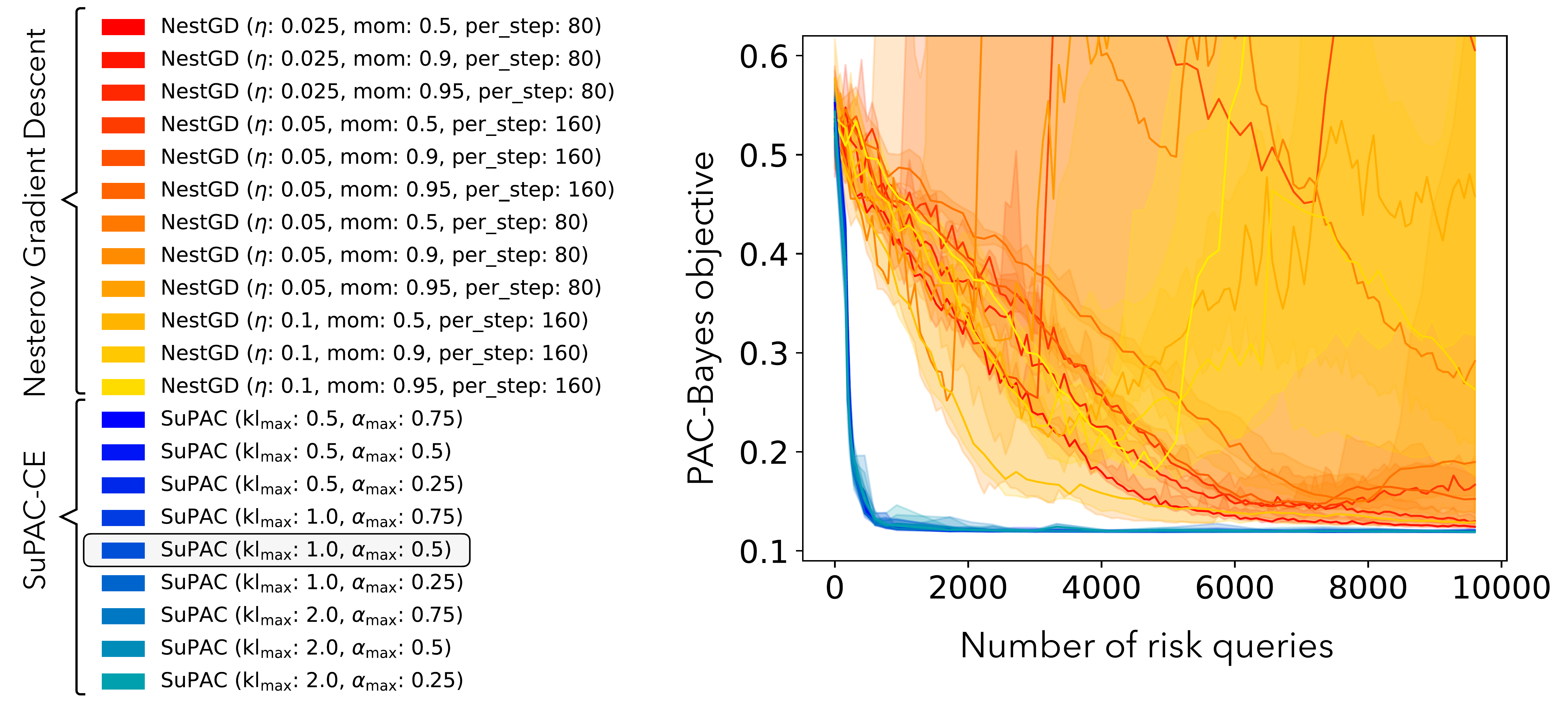}
\label{fig:nesterov_vs_SuPAC_tot}
\caption{Comparison of \algo with Nesterov accelerated gradient descent for a variety of hyperparameters choices. Each optimisation procedure was repeated 8 times; the median performance and 0.2 and 0.8 quantiles are represented.  \algo proved to be consistently more efficient for all hyperparameters values tested. The hyperparameter for \algo assessed in the main part of the publication is highlighted.}
\end{figure}

\algo was further compared to Nesterov accelerated gradient descent (implementation in the publication repo). Starting from the two sets of hyperpara\-meters preselected for GD, optimisation procedures using a momentum of 0.5, 0.9 and 0.95, and either the original step size or twice the step size were assessed. Each of these 12 new optimisation procedures was repeated 8 times, and compared to \algo (see \Cref{fig:nesterov}). For no choice of hyperparameter values did Nesterov accelerated GD proved more efficient than \algo (\Cref{fig:nesterov_vs_SuPAC_tot}). The increase of step size in conjunction with the moderate momentum improved the speed of the optimisation procedure, but at the cost of a higher risk of optimisation failure, leading to 3 out of 8 runs (resp. 2 out of 8 runs) for 160 simulations per step (resp. 80 simulations per step) with a final objective higher than the initial objective. Higher momentum led to major instabilities, with less than 3 runs out of 8 managing to reduce the objective below 0.2 (compared to 0.121 obtained by \algo) for all hyperparameter combinations. For the original step size, momentum appeared to improve the stability of the procedures for all setting except moderate momentum for a per step hyperparameter of 80. Higher momentum procedures led to a speed decrease, caused by the larger number of steps necessary for momentum to build up.

\begin{figure}
\centering
\includegraphics[width=.95\textwidth]{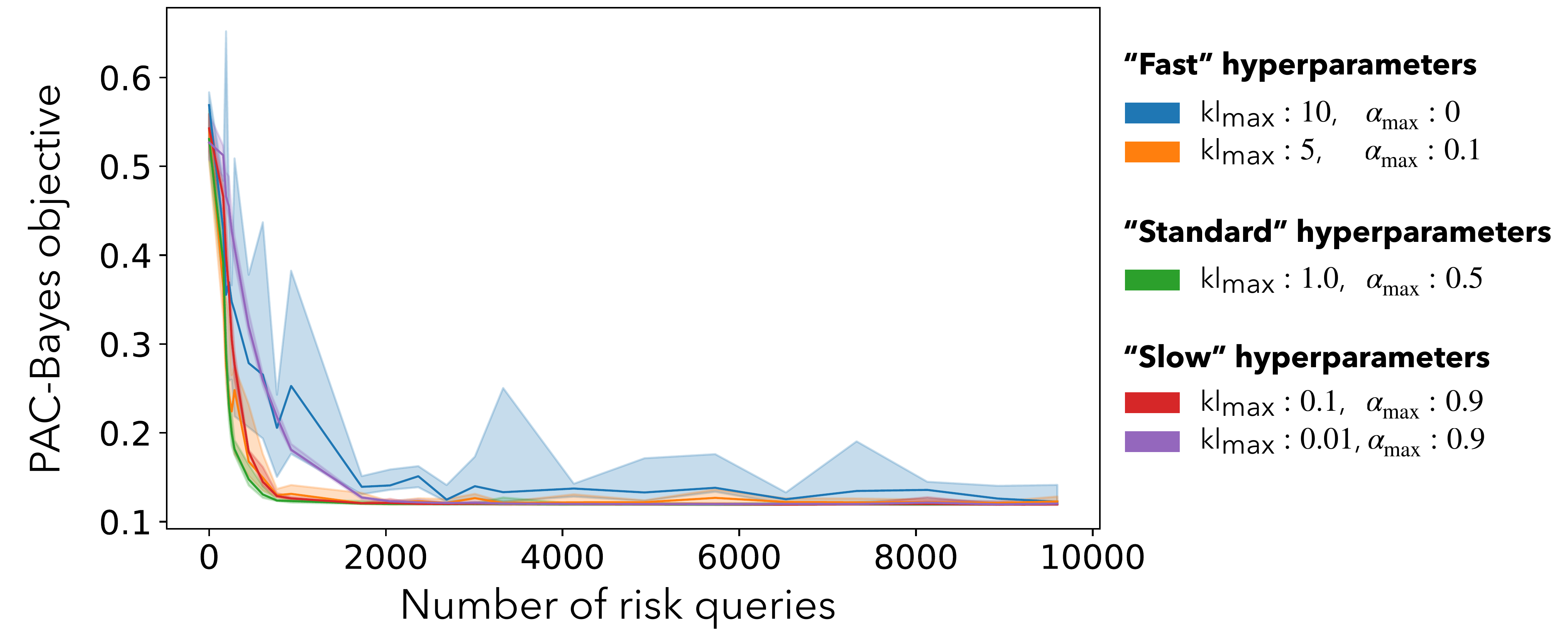}
\label{fig:supac_ext_hyperparams}
\caption{Performance of \algo with extreme hyperparameters values. Each optimisation procedure was repeated 8 times; the median performance and 0.2 and 0.8 quantiles are represented. \algo exhibited noticeable instabilities and speed loss for hyperparameters leading to insufficient regularization (blue curve). Too much regularisation lead to speed decrease in the early phase of the optimisation procedure (purple curve)}
\end{figure}

The impact of \algo's hyperparameters was investigated by running further optimisation procedures with different choices of hyperparameters. A grid was assessed, with values of $\klmax$ in $\{0.5, 1, 2\}$ and $\dampen$ in $\{0.25, 0.5, 0.75\}$, with each optimisation process repeated ten times (see \Cref{fig:nesterov_vs_SuPAC_tot}). The resulting optimisation procedures proved to all have similar performances, with only a slight decrease in speed in the early phase between the most regularized and less regularized hyperparameters which was below the noise level after the fourth optimisation step (see \Cref{fig:nesterov_vs_SuPAC_tot}). Two further sets of slow hyperparameters values ($(\klmax, \dampen) \in \{(0.1, 0.9), (0.01, 0.9)\}$) and fast hyperparameters values ($(\klmax=5, \dampen=0.1)$, $(\klmax=10, \dampen=0)$) were also assessed, with 8 repeats (see \Cref{fig:supac_ext_hyperparams}). The slow hyperparameters led to more stable and reproducible optimisation procedures. For the small maximum step size of $\klmax=0.01$, the average performance of the optimisation process was similar (\emph{i.e.} difference below the noise level) to the performance of the optimisation process with standard hyperparameters after 2000 risk queries. The highest maximal step size assessed of $\klmax=10$ resulted in a final average PAC-Bayes bound of $0.147 \pm 0.022$, with a standard deviation between runs of $0.061$, significantly higher than the standard deviation for the standard hyperparameters ($0.0032$, p-value of $1.95e-09$).

Computations were performed using Azure Machine Learning compute clusters with 32 cores and Intel Xeon Platinum 8272CL processors.

\subsection{Meta-Learning experiments}
For the meta-learning experiments, the tasks were generated as follow. Empirical risk functions of form
\begin{equation}
R_{\omega, A, x_0}:x\mapsto\tanh(h(\omega \lVert A( x - x_0)\rVert^2)/10)
\end{equation}
with $h(x)= \cos(x) +x $ were considered. These are such that $x_0$ is the only global minima of $R_{\omega, A, x_0}$, while all $x$s such that $\omega \lVert A( x - x_0)\rVert^2 = \pi/2 + 2k\pi$ are local minima. The distributions of the risk parameters are as follow: $x_0\sim \mathcal{N}(\tilde{x}_0, \Sigma_0)$, $\omega\sim \mathcal{U}(\frac{3}{2}\pi, \frac{5}{2}\pi)$ and $A_{i,j}\sim \mathcal{N}(\delta_{i,j}, \sigma^2=0.05^2)$. The mean parameter $\tilde{x}_0$ was initiated at random on the sphere of radius $2$, while the covariance $\Sigma_0$ was initiated at random as
\begin{equation*}
\Sigma_0 = O \times\text{diag}(\sigma_1^2, \dots, \sigma_d^2)\times O^t, 
\end{equation*}
where $\sigma_1, \dots, \sigma_{d-2} = 0.05$, $\sigma_{d-1}, \sigma_{d} \sim \exp(\mathcal{U}(-0.5, 0.5))$ and $O$ is drawn at random amongst orthonormal matrices. The dimension of the predictor space $d$ is fixed to $8$.

The meta training process was performed as follow. The initial calibration phase for each task was performed in 15 steps, with 100 score queries for the first five steps and 50 score queries for the remaining steps. The hyperparameters were set to $\klmax = 0.5$, $\dampen = 0.3$ and $10^4$ samples are used to estimate weights. This initial meta step used a mini batch size of 10, a maximum meta $\text{kl}$ step of $0.2$ and step size of $\lambda^{-1}$.
After all tasks have been trained once, the hyperparameters for \algo were modified: the number of steps was reduced to $4$, and $\dampen$ set to $0.7$. 20 risk queries are performed on the first and third step, and none on the second and fourth. This accounts for the fact that the posterior distribution updates are expected to be small at this stage. The mini batch size is increased to 20. After 19 epochs, the step size is reduced to $0.5 \lambda^{-1}$ and the maximum meta $\text{kl}$ step to $0.1$. After 30 more epochs, the step size was reduced to $0.4 \lambda^{-1}$, and trained for a further 100 epochs.

The performance of sequence of priors was assessed in the following way. 40 test tasks were drawn. For each prior, a full independent calibration was performed on each task, using 20 steps of \algo (100 risk queries for the first 5 steps, 50 for the remaining steps). The resulting posterior performance is assessed by computing the bound using $10^4$ fresh evaluations of the risk. The mean of these performance over the task defines the meta test score. The dispersion of these test performance between different test task is assessed by computing the quantiles 0.2 and 0.8 of the test performances at a given prior. This procedure being quite computationally intensive, only the first ten priors constructed and afterwards one prior out of five were assessed.

Computations were performed using Azure Machine Learning compute clusters with 16 cores and Intel Xeon Platinum 8272CL processors.

\end{document}